\documentclass{egpubl}

\usepackage[T1]{fontenc}
\usepackage{dfadobe}  

\usepackage{cite}  
\BibtexOrBiblatex

\electronicVersion
\PrintedOrElectronic

\ifpdf \usepackage[pdftex]{graphicx}
\else \usepackage[dvips]{graphicx} \fi

\usepackage{amsmath}
\usepackage{siunitx}
\usepackage{amsbsy}

\newcommand{\unit}[1]{\, \mathrm{#1}}

\newcommand{\mat}[1]{\mathbf{#1}}

\renewcommand{\vec}[1]{\mathbf{#1}}

\newcommand{\norm}[1]{\left\lVert #1 \right\rVert}

\newcommand{\transpose}[1]{{#1}^{\! \top}}

\newcommand{\summation}[3]{\ensuremath{\sum_{#2}^{#3} #1}}

\DeclareMathOperator{\sign}{sign}

\title[Image-Based Alignment of 3D Scans]%
{Image-Based Alignment of 3D Scans}

\author[D. Messer et al.]
{\parbox{\textwidth}{\centering Dolores Messer$^{1}$,
        Jakob Wilm$^{1,2}$,
		Eythor\,R. Eiriksson$^1$,
		Vedrana\,A. Dahl$^1$,
		and Anders\,B. Dahl$^1$
        }
        \\
{\parbox{\textwidth}{\centering $^1$Technical University of Denmark, Visual Computing, Denmark\\
         $^2$ University of Southern Denmark, Maersk Mc-Kinney Moller Institute, Denmark
       }
}
}

\begin{document}
	
	\maketitle
	\begin{abstract}
		Full 3D scanning can efficiently be obtained using structured light scanning combined with a rotation stage. In this setting it is, 
		however, necessary to reposition the object and scan it in different poses in order to cover the entire object. In this case, 
		correspondence between the scans is lost, since the object was moved. In this paper, we propose a fully 
		automatic method for aligning the scans of an object in two different poses. This is done by matching 2D features 
		between images from two poses and utilizing correspondence between the images and the scanned point clouds. 
		To demonstrate the approach, we present the results of scanning three dissimilar objects.

		\ccsdesc[300]{Computing methodologies~3D imaging}
		\ccsdesc[500]{Computing methodologies~Matching}
		\ccsdesc[100]{Hardware~Scanners}		

        \printccsdesc
        
        \vspace{0.15cm}
        {\normalfont \textbf{Author keywords}}\\
        SIFT 2D features; feature matching; RANSAC; structured light 3D scanner; cultural heritage.
        
	\end{abstract}  
	
	\section{Introduction}
	
	Surface scanning in 3D has been used for a long time for metrology in industrial applications, and this has expanded to new application areas \cite{sansoni2009state}. A case of both scientific and educational value is found in digital preservation and analysis of cultural heritage 
	~\cite{levoy2000digital,anderson2002unwrapping}, e.g. for scanning specimens such as skulls. When digitizing collections of for example hundreds of specimens, it is important to have an efficient and uncomplicated protocol.
	
	\begin{figure*}[htb]
		\centering
		\frame{\includegraphics[width=.7\linewidth]{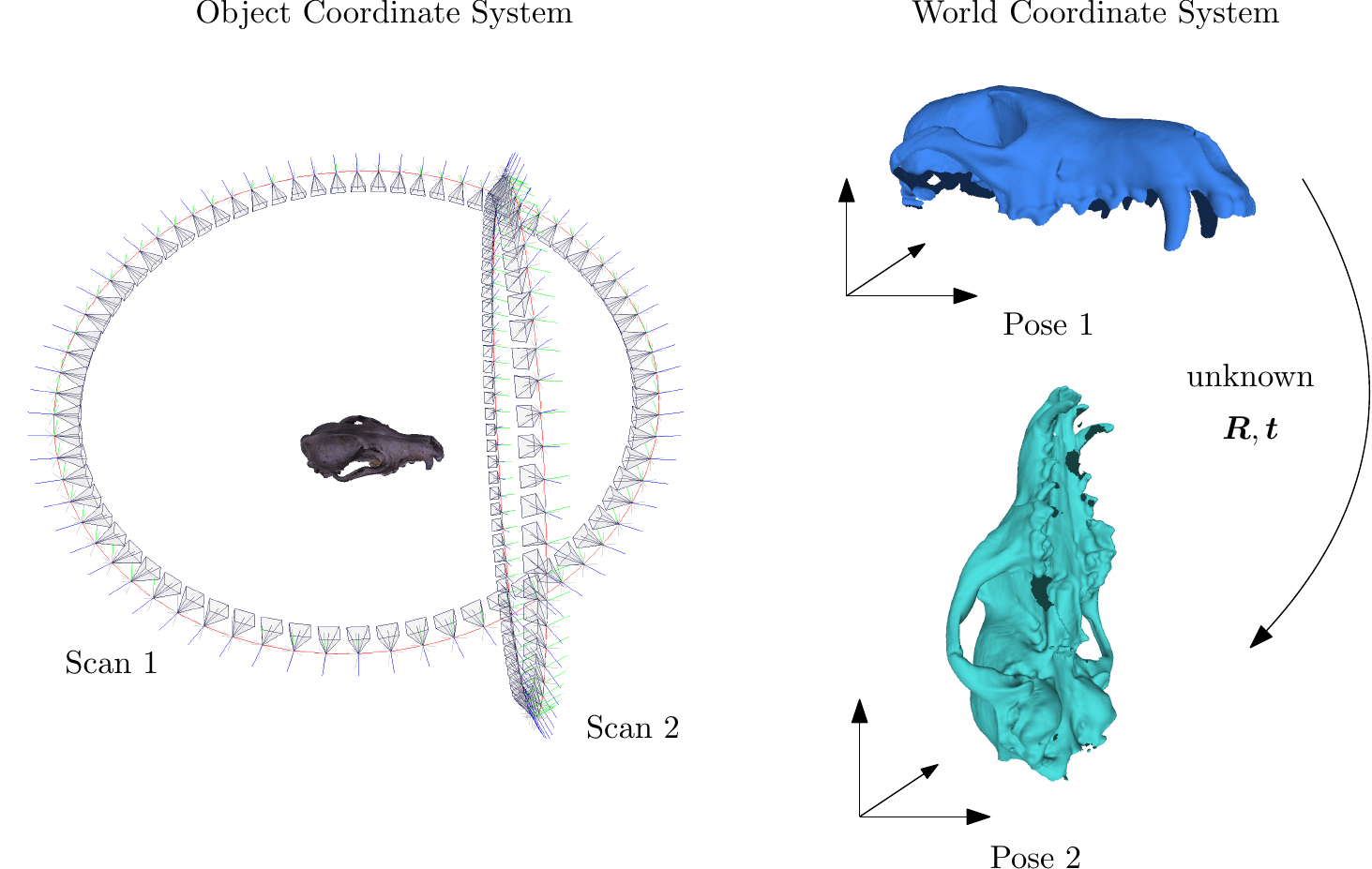}}
		\caption{\label{fig:camera_rings_alignment_problem} When an object is scanned on a rotation stage and in two different poses, the camera positions in the object coordinate system define two circles, typically with two intersecting views. The spatial relationship between the two circles is unknown, as it depends on the pose of the object. In the world coordinate system, two resulting scans are related by the same unknown rigid transformation. This paper describes a method to recover rotation $\boldsymbol{R}$ and translation $\boldsymbol{t}$, and align scans from different poses. We utilize 2D camera images obtained during scanning.}
	\end{figure*}
	
	The work presented here is motivated by digitizing archaeological and zoological specimens for morphological analysis using a structured light scanner, where high accuracy point clouds of the objects are acquired \cite{niven2009virtual,trinderup2016}. Our overall ambition is to design an off-shelf, low-cost and easy-to-use system which is based on an existing structured light scanner \cite{wilm2016,eiriksson2016precision}. Any person working at a museum should be able to use our system in order to obtain a full scan of an archaeological object.
	
	There are real-time scanning systems such as handheld 3D scanners available in order to scan small objects. However, Kersten et al.~\cite{kersten2016} were testing the geometrical accuracy for different handheld 3D scanners, and demonstrated that the evaluated handheld 3D scanners do not reach the quality and accuracy of structured light systems.
	
	Using a rotation stage in the scanning process has the advantage that sub-scans can be acquired in regular angular intervals. Since the geometry of the rotation stage is known, these sub-scans are combined with very high accuracy. A 360 degree scan will, however, typically not cover areas that are occluded, like the bottom of the object \cite{trinderup2016}. Therefore, the object needs to be scanned in different poses. Based on our experience, it takes typically 3--4 poses to cover a whole archaeological object, such as for example a skull. In general, when the pose of an object is changed, all known correspondence is lost, and a new partial scan is initiated. Fig.~\ref{fig:camera_rings_alignment_problem} illustrates the problem of changing pose when scanning surfaces of the same object from two sides. Note that some parts of the surface are only visible in one of the scans.
	
	Automatically aligning two or more scans from different poses requires matching partial scans that are typically far apart. In this paper, we present a fully automatic method for aligning two point clouds scanned using a structured light 3D scanner. Our method is utilizing the raw 2D images recorded by the scanning system. Thus, we are taking advantage of all available captured data. Using well established and robust methods for 2D point correspondence matching, we can derive 3D point correspondences between point clouds obtained from different poses, because correspondence between the images and the 3D point clouds is given by the scanner. The pipeline presented can serve as a robust global registration technique, or as a pre-processing step for further registration refinement algorithms.
	
	\subsection{Related Work}
	\label{sec:related_work}
	Markers for guiding the registration is one solution for robustly aligning 3D scanned data. However, markers will become part of the scan and leave undesired data in its texture and geometry. The problem can be solved by manually annotating point correspondences between the two sets~\cite{niven2009virtual}. For scanning a large number of objects, this manual annotation quickly becomes a cumbersome task.
	
	Iterative Closest Point (ICP)~\cite{besl1992method,Chen1992} is commonly used for registering point clouds that are relatively well aligned. The main idea behind the ICP algorithm is to iteratively minimize a cost function which describes the difference between the two considered 3D point clouds. It is, however, not global in the sense that good pre-alignment is generally necessary for the method to converge to the global minimum. In order to achieve global alignment, feature correspondences between datasets need to be obtained.
	
	Bendels et al.~\cite{bendels2004image} proposed an image based registration method for aligning point clouds from laser range scanning, a related field of research. They use the Scale-Invariant Feature Transform (SIFT)~\cite{lowe2004distinctive} to obtain point correspondences between several scans. A RANSAC (Random Sampling Consensus) scheme~\cite{fischler1981random} is used for outlier removal and domain-constrained ICP variant used for alignment. Finally, the global registration error is minimized via a graph relaxation technique.\\ 
	Similarly, B{\"o}hm and Becker~\cite{bohm2007automatic} again used SIFT features and a RANSAC filtering scheme in the registration of terrestrial laser scans. They computed a rigid body transform between scans and the results were comparable to manually annotated results.\\
	There are several other approaches for registration of point clouds, which are obtained with terrestrial laser scanners, involving feature extraction from images~\cite{Seo2005,Wang2008,Kang2009,Alba2011,Barnea2012}. Weinmann~\cite{Weinmann:2016} provides a detailed overview over point cloud registration of data that is acquired with terrestrial laser scanners and range data.
	
	As an alternative to image-based features, three dimensional shape descriptors can be used to determine point correspondences~\cite{tam2013registration,haensch2014comparison,Theiler2014}. However, they most often do not exploit texture information, which is available when using image based scanning techniques such as structured light scanning. 

    Weinmann and Jutzi~\cite{Weinmann2011} propose to use the number of feature correspondences between respective images in order to measure the similarity of different scans. They use this approach for automatically organize a given number of unorganized scans, and for successive pair-wise registration.

	\begin{figure*}[ht!]
		\centering
		\includegraphics[width=.8\linewidth]{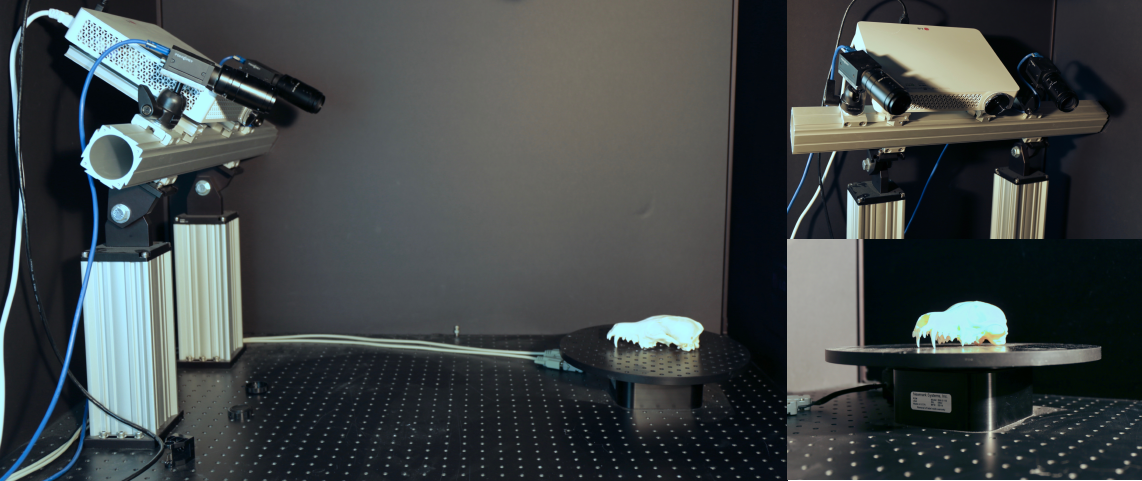}
		\caption{\label{fig:hardware0} Our scanning setup, which is a stereo structured light system with a calibrated rotation stage outputting aligned 360 degree scans: full setup (\emph{left}), stereo camera with the LED projector (\emph{right top}), and the rotation stage (\emph{right bottom}).}
	\end{figure*}

	In line with the above mentioned techniques, our alignment method uses SIFT feature descriptors applied on affine invariant interest points~\cite{Mikolajczyk2010} from 2D images obtained during scanning. Similarly to Weinmann and Jutzi~\cite{Weinmann2011}, we propose to exploit the number of matched features between the image sequences of the different poses in order to measure similarity of different part-scans. We use this approach to find the views with the largest overlap, and to subsequently estimate the transformation between the two poses based on these views. Our approach has the following advantages:
	\begin{itemize}
		\item The entire processing pipeline is fully automatic and fast compared to manual alignment.
		\item 2D image features are well investigated and easy to visualize, which makes it easy for a user at a museum to assess the quality of matches.
		\item High quality texture information is used, so our method can also handle objects with a less distinct 3D geometry.
	\end{itemize}
	
	\section{Method}
    We start by explaining the existing single-pose data acquisition pipeline and then describe our two-pose alignment technique. The datasets obtained by the single-pose acquisition are the input to our alignment method. 
	
	\subsection{Single-pose acquisition}
	Our data is acquired on a stereo structured light setup consisting of two industrial cameras and a high resolution LED digital projector. The object is placed on a microrotation stage which has 70 arc-second accuracy according to manufacturer specifications. 
	Fig.~\ref{fig:hardware0} shows our scanner. We perform stereo calibration~\cite{Zhang1999a}, and calibrate the rotation stage axis~\cite{Chen2014a}.
	
	We employ a phase shifting method for scene encoding, where correspondences are found by matching the phase maps of shifting sinusoidal patterns. The sinusoidal wavelength is approximately $1$ cm on the object surface. In order to recover phase ambiguities we employ phase unwrapping based on the \emph{heterodyne} principle~\cite{Reich1997a}. This yields highly detailed point clouds with a homogeneous point spacing of approximately $0.21$ mm on the object surface. The absolute accuracy and precision of this setup is comparable to commercial metrology grade scanners, as demonstrated in~\cite{eiriksson2016precision}.
	
	The rotation stage allows us to acquire scan sequences in equal angular steps. Due to pre-calibration, those are fully aligned, resulting in a single point cloud per pose. The angular interval can be chosen freely. In our experiments, we used $5 \unit{degree}$ intervals, resulting in $72$ part-scans per sequence. In addition, all underlying data may be retrieved. Most importantly, this includes the two fully lit images from the stereo setup and corresponding point clouds from each angle.
	
	\subsection{Two-pose alignment}
\begin{figure*}[htb]
		\centering
		\includegraphics[width=.8\linewidth]{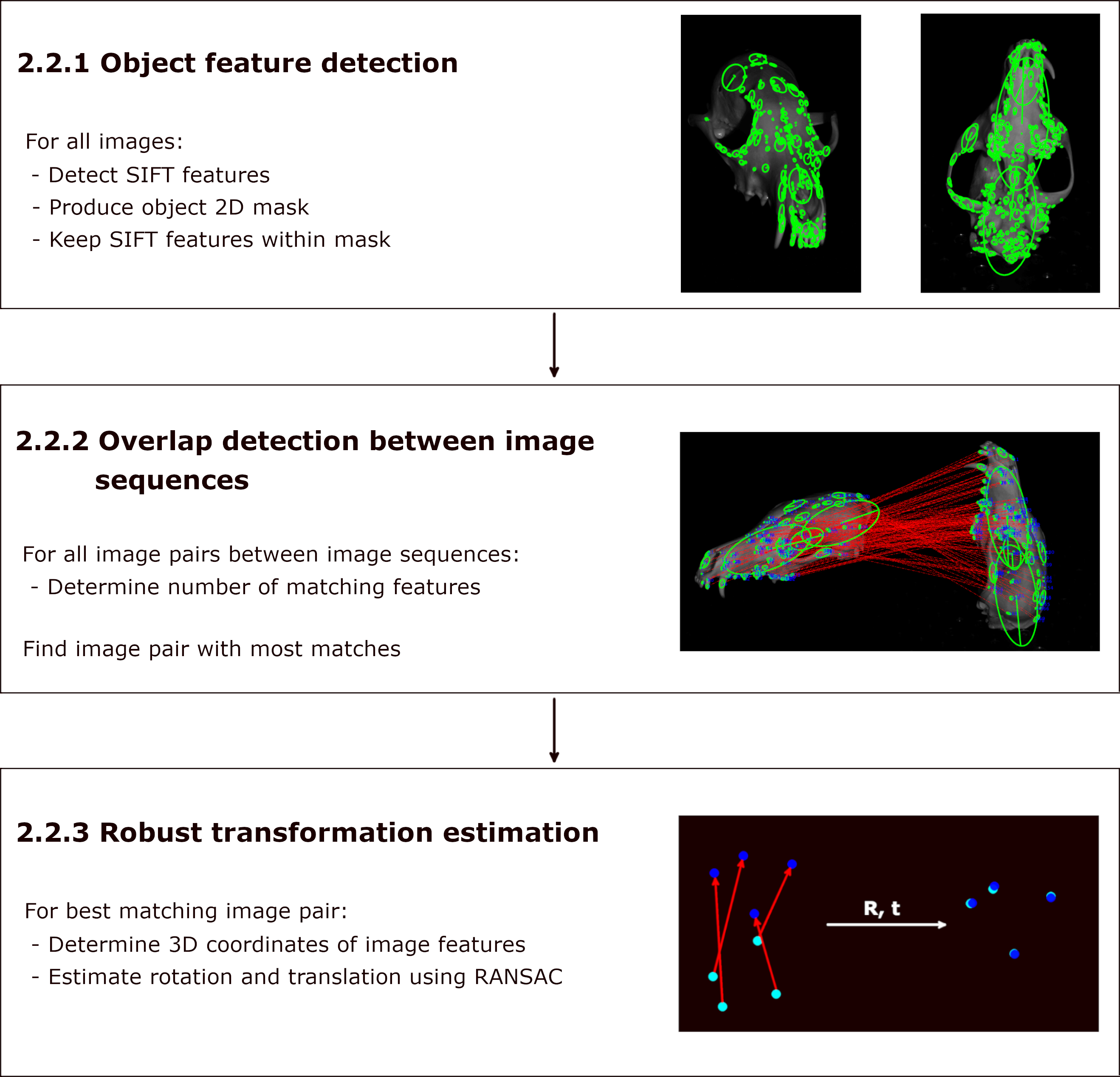}
		\caption{\label{fig:flowchart_IPE} Flowchart of our proposed method.}
	\end{figure*}

	Our aim is to align two point clouds, $P$ and $Q$, which represent two different poses of the same object, by using the two image sequences obtained from scanning. To do so, we need correspondences between three or more points from $P$ and $Q$. Our idea is to find one image from one pose and another image from the other pose, where the object is viewed from the same side. We can then use image features to establish correspondence between image pixels. Using the correspondences between 2D images and 3D point clouds as given by the scanner, this will lead to 3D correspondences. The complete method is outlined in Fig.~\ref{fig:flowchart_IPE}.

	We need to address two problems. First, we must choose images with high overlap from the two sequences. Second, we need to compute the 3D transformation utilizing 2D correspondences in a robust manner. Both these problems are solved using SIFT.
	
	\subsubsection{Object feature detection}
	We use fully lit images acquired as part of the structured light pattern sequence obtained at each angle. Based on these images, we compute SIFT including its descriptor and interest point, i.e. the position in the image. Due to known geometry, we have a correspondence between SIFT features and the scanned object. Hereby, we can select the SIFT interest points on the object and discard those from the background. Similarly, we obtain the 3D coordinates of the SIFT features as follows: First, the 3D part-scan points are projected into the corresponding images, then the interest points are matched with their respective nearest projected point, and the final 3D coordinates associated with the SIFT descriptors are given by the original 3D coordinates of the nearest projected points. To make our method more robust to potential large viewpoint changes due to different angles between images of two poses, we compute descriptors that are adapted to allow for an affine transformation between views~\cite{Mikolajczyk2010}. In order to compare the results obtained by using our method on different objects, a threshold was manually chosen for each object, such that the total number of SIFT feature descriptors per image is around $1000$.
	
	\subsubsection{Overlap detection between image sequences}
	During scanning, we obtained two image sequences from different poses. We want to find an image from the first sequence and an image from the second sequence, where the object is seen from the same angle, see Fig.~\ref{fig:camera_rings_alignment_problem}. Note that the camera 
	positions define two circles relative to the object and those circles typically share two views. The aim is to find views with a large overlap of the seen object. For this purpose, we consider pairwise matching of SIFT features between all images of the two sequences.
	
	In our current implementation, we use the pair of images with the highest number of matched features for estimating the transformation. On these two images, the object is seen from a similar direction, hence both the illumination condition and the SIFT descriptors are similar for these two images. To dismiss faulty feature matches, we first employ the matching criterion of Lowe~\cite{lowe2004distinctive}. Specifically, features are only 
	matched if the ratio between the best matched descriptor and the second best is less than 0.5.
	
	\begin{figure*}[ht!]
		\centering
		\begin{tabular}{c}
			\includegraphics[width=.75\linewidth]{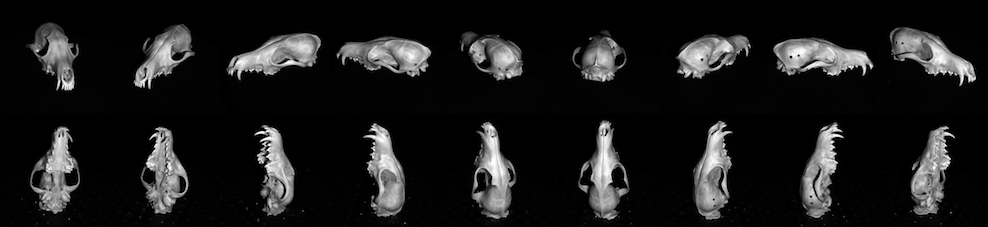}\\
			\textit{(a) Fox skull} \\
			\includegraphics[width=.75\linewidth]{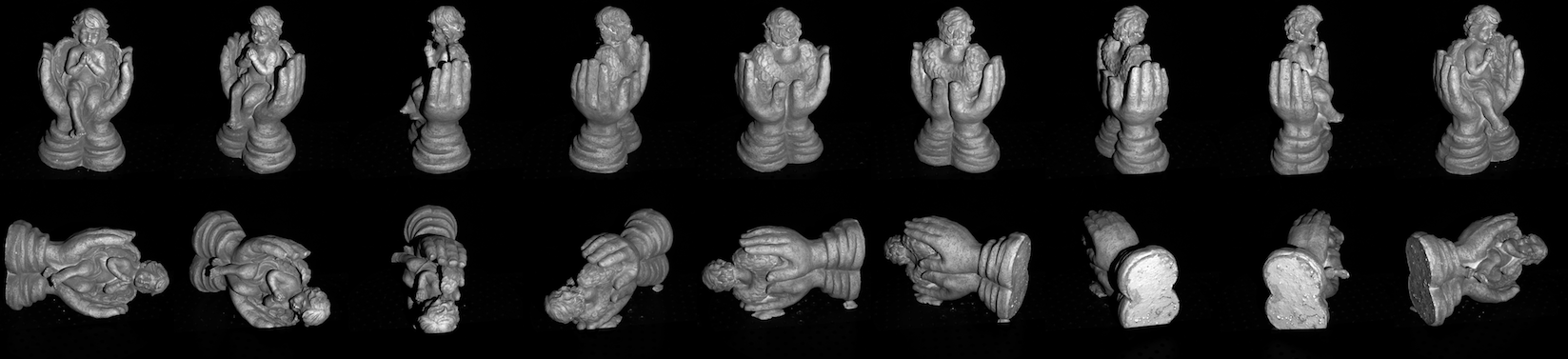}\\
			\textit{(b) Angel} \\
			\includegraphics[width=.75\linewidth]{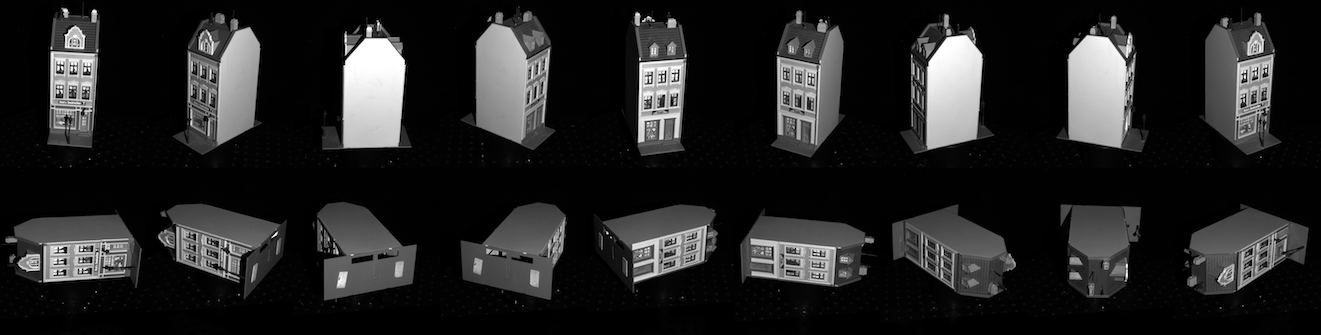}\\
			\textit{(c) House}
		\end{tabular}
		\caption{\label{fig:images} Image sequences of two different poses for the three objects. For better visualization, only images taken after the rotation stage has turned by a multiple of 40 degrees are shown.}
	\end{figure*}
	
	\subsubsection{Robust transformation estimation}
	\label{sec:transformation_estimation}
	From the 3D points associated with each SIFT feature and the matching SIFT descriptors, we now obtain candidates for 3D point correspondences between the point clouds of the two poses. The matching descriptors are not guaranteed to be correct, so to obtain robust estimation of the 
	transformation, we employ the RANSAC heuristic and sample 4 correspondences. Based on these, a rigid transformation is estimated. 3D points which are transformed to within $1.0$~$\unit{mm}$ of their correspondence point are defined as inliers. This procedure effectively exploits the 
	discriminative power of the 3D geometry while working with simple and fast 2D features.
	
	Estimating the transformation between coordinate systems using the sets of corresponding 3D points is known as the absolute orientation problem. Specifically, given point correspondences $(\vec{p}_i, \vec{q}_i)$, the problem amounts to finding rotation $\mat{R}$ and translation 
	$\vec{t}$ which minimizes
	\[
	\summation{\norm{ \left( \mat{R} \vec{p}_i + \vec{t} \right) - \vec{q}_i }^2}{i}{} \quad .
	\]
	
	The proposed solutions vary only slightly in accuracy, robustness, stability and efficiency \cite{eggert1997estimating}. Hence, we use the classical Kabsch algorithm~\cite{Kabsch1976}. Given covariance matrix
	\[
	\mat{C} = \summation{\left(\vec{p}_i - \bar{\vec{p}} \right) \transpose{\left(\vec{q}_i - \bar{\vec{q}} \right)}}{i}{} \quad ,
	\]
	where $\bar{\vec{p}}, \bar{\vec{q}}$ are the respective centroids, the optimal rotation is found from the SVD, $\mat{C} = \mat{U} \boldsymbol{\Sigma} \transpose{\mat{V}}$, as
	\[
	\mat{R} = \mat{U} \begin{bmatrix} 1 & 0 & 0 \\ 0 & 1 & 0 \\ 0 & 0 & d \end{bmatrix} \transpose{\mat{V}} \quad , \quad d = \sign(\det(\mat{V} \transpose{\mat{U}})) \quad ,
	\]
	while the optimal translation is
	\[
	\vec{t} = \mat{R} \bar{\vec{p}} - \bar{\vec{q}} \quad .
	\]
	
	The Kabsch algorithm is run for each iteration of RANSAC, and finally based on all inliers of the most successful iteration. This transformation of partial scans is then followed by a few iterations of ICP for fine alignment. We use the ICP variant suggested by Rusinkiewicz and Levoy \cite{Rus2001,Pulli1999}, where the transformation is computed based on 75\% of the closest nearest neighbors.

	\section{Results}
	For testing our method, we acquired data sets from three case objects which exhibit different characteristics. The first object, a fox skull, is chosen due to the scanner's application for digitizing archaeological and zoological specimens. The other two objects are chosen to test 
	the performance of our method on an object with well defined texture, and an object with a repetitive pattern.
	
	Each object was scanned in two different poses at an angular resolution of 5 degrees. A sample of the resulting fully lit images are shown in Fig.~\ref{fig:images}.

	\begin{figure*}[ht]
		\centering
		\begin{tabular}{c c c}
			\includegraphics[width=.3\linewidth]{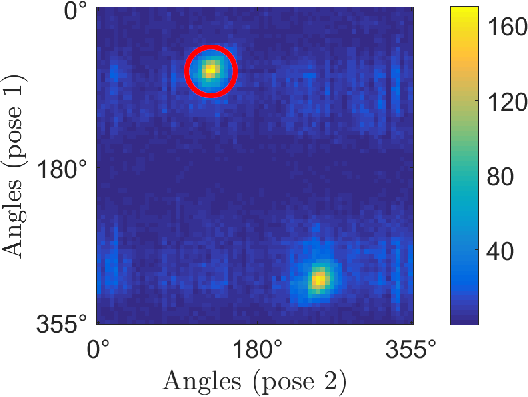} & \includegraphics[width=.3\linewidth]{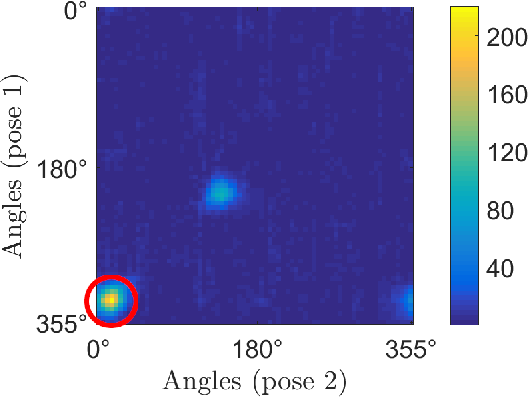} & \includegraphics[width=.3\linewidth]{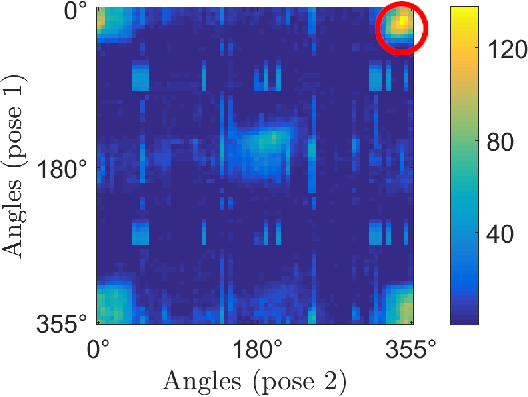} \\
			\textit{(a) Fox skull} & \textit{(b) Angel} & \textit{(c) House}
		\end{tabular}
		\caption{\label{fig:matches} Number of SIFT feature matches. In all three cases, two distinct maxima occur, corresponding to the intersections of circles depicted in Fig.~\ref{fig:camera_rings_alignment_problem}. The red circles highlight the image pairs with most matches. Note that angles wrap in both directions. Thus, in case of the house, 
			high-match regions in the corners correspond to a single matching region.}
	\end{figure*}

	\begin{figure*}[ht]
		\centering
		\begin{tabular}{c c c}
			\includegraphics[width=.3\linewidth]{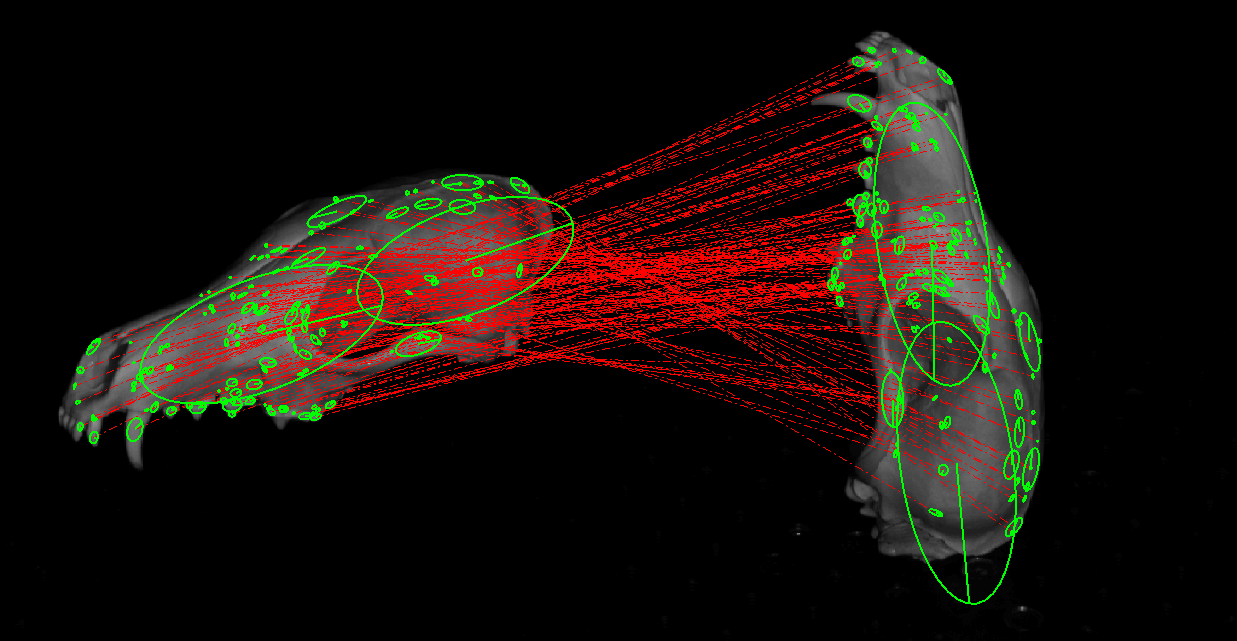} & \includegraphics[width=.3\linewidth]{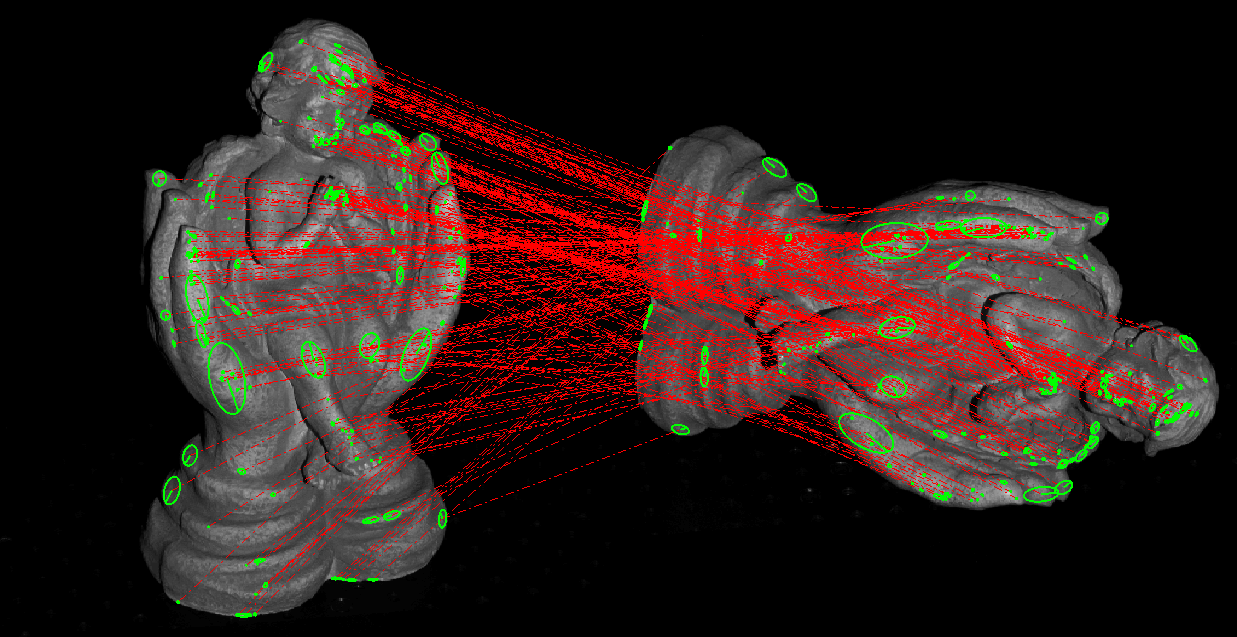} & \includegraphics[width=.3\linewidth]{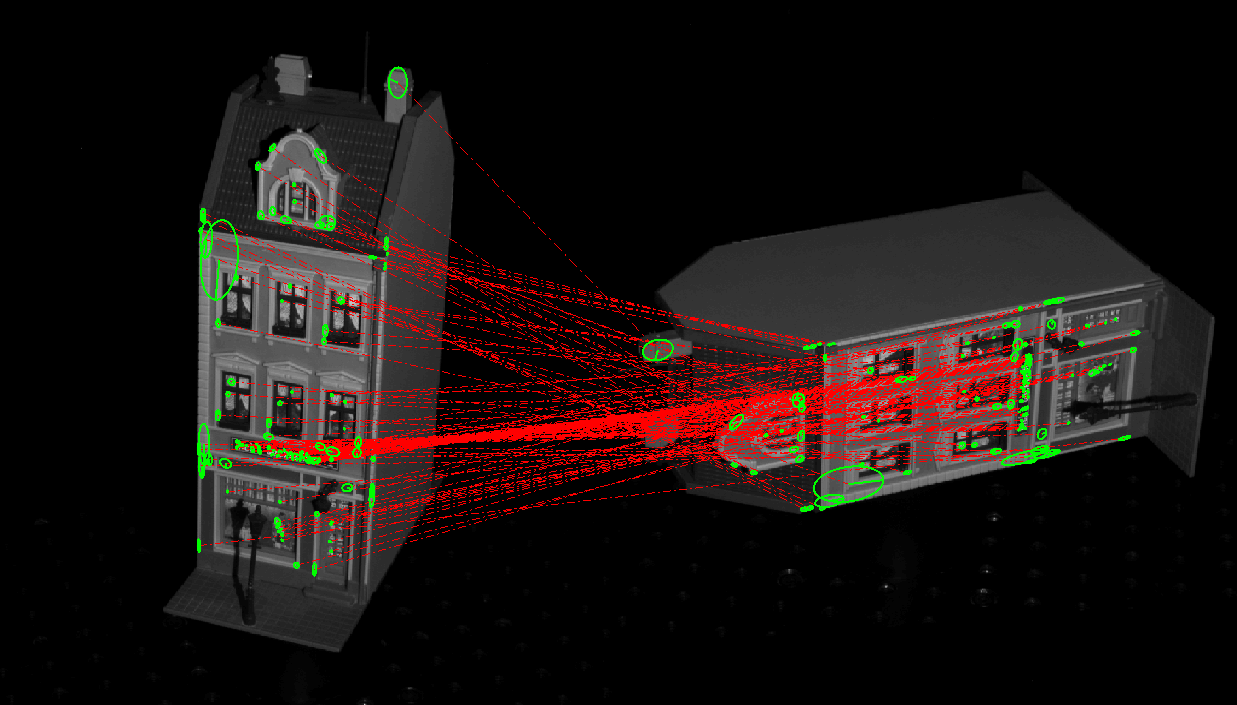} \\
			\textit{(a) Fox skull} & \textit{(b) Angel} & \textit{(c) House}
		\end{tabular}
		\caption{\label{fig:SIFT_bestMatch} RANSAC inliers of best SIFT feature match images. The size and the orientation of the ellipse corresponds to the scale and the orientation of the feature.} 
	\end{figure*}
	
	\subsection{Feature matching}
	Fig.~\ref{fig:matches} shows the resulting number of matches for each image pair for the three case objects. We can see that a relatively high angular resolution of 5 degrees is needed in order to find the maximum. Furthermore, we obtain two clear maxima corresponding to the 
	intersections of scan circles depicted in Fig.~\ref{fig:camera_rings_alignment_problem}. The red circles highlight the image pairs with most matches. In case of (a) the fox skull, the results indicate that both maxima might be equally good. This is due to the skull appearance which has a stable number of features independent from the side it 
	is viewed from. However, both (b) the angel and (c) the house have a clear best matching image pair. For both objects, the best image pair is when the front side of the objects is viewed, since this side contains many features. The back side of those objects contains less features, 
	so the second maximum is less good. Moreover, the house exhibits several local maxima which can be explained by the repetitive elements.
	
	Fig.~\ref{fig:SIFT_bestMatch} shows the matched SIFT feature descriptors for the image pair with most matches, i.e. the image pair that corresponds to the global maxima in Fig.~\ref{fig:matches}. Only the RANSAC inliers are shown. Table~\ref{tab:results} lists the ratio of RANSAC 
	inliers versus the total number of matches for these best matching pairs. It can be seen that the vast majority of matches were kept for estimation of the transformation matrix.
	
	\subsection{Final Transformation}
\begin{figure*}[htb]
		\centering
		\frame{\includegraphics[width=.8\linewidth]{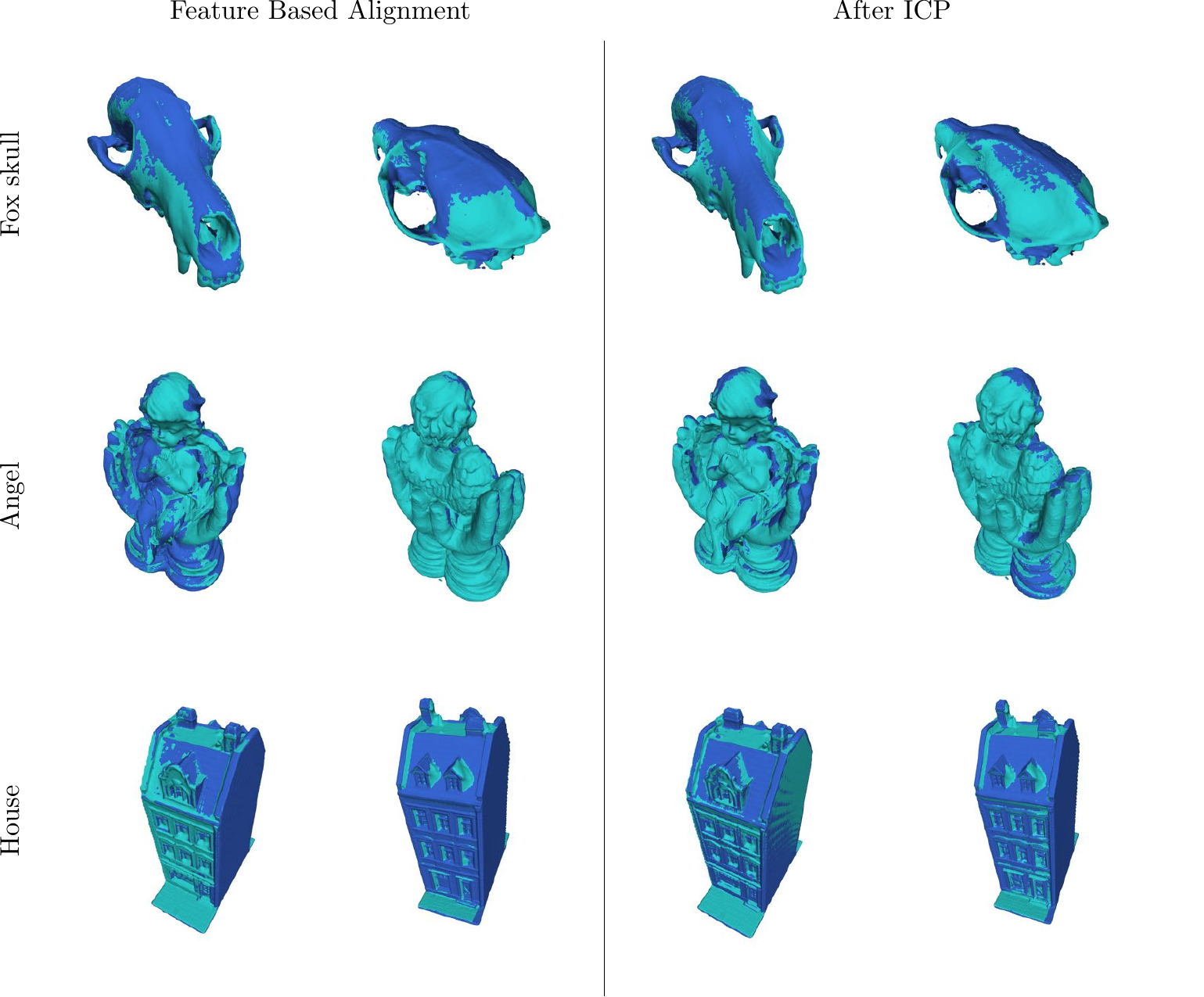}}
		\caption{\label{fig:result_matrix} Alignment results obtained with our method. The objects are shown from the front and back side. For the fox skull and angel data, ICP slightly improves the alignment result, while for the house, it provides a clear improvement.}
	\end{figure*}

	Fig.~\ref{fig:result_matrix} presents the final result of our image-based alignment both before and after ICP for all objects seen from two different angles. ICP is only marginally improving the alignment obtained with our algorithm. This finding is confirmed by the movement after 
	ICP, which is in the order of $1~\unit{mm}$ for all objects (see Table~\ref{tab:results}). The overall computation time is in the order of $10~\unit{minutes}$, which is sufficient for our needs.

	\begin{table}[!h]
		\centering
		\resizebox{\linewidth}{!}{
			\begin{tabular}{l|c|c}
				Object & RANSAC Inliers/All matches & Movement after ICP (RMSE) \\
				\hline
				Fox skull & $158/170$ = $0.93$ & $0.38$~\si{mm}\\
				Angel & $213/220$ = $0.97$ & $1.19$~\si{mm} \\
				House & $127/138$ = $0.92$ & $0.35$~\si{mm}
		\end{tabular}}
		\caption{Best number of RANSAC inliers and the RMSE of point movements obtained with ICP after our method. Fig.~\ref{fig:result_matrix} shows visual results.}\label{tab:results}
	\end{table}
	
	\section{Discussion and conclusion}
	We have presented a fully automatic method for alignment of part scans, a task encountered in many 3D scan applications. The method is based on well established 2D image features and is computationally inexpensive. It successfully aligns three very different objects. Despite we have only used the method to align two poses, adding a new pose is done by matching to one of the other and making the final adjustment using ICP to the collected point cloud.
	
	The insight provided by our work is twofold. First, we show that the features matching maxima depicted in Fig.~\ref{fig:matches} have a width corresponding to approximately $20$ degrees. Thus, in order to ensure that suitable data is available, images need to be acquired from around 
	$18$ angular scan positions. This, however, is only necessary for fully lit images, not for 3D data. As a result, the method can be employed in many scenarios with only small adaptations to the scan protocol.
	
	Secondly, the accuracy of the automatic alignment procedure is high. As ICP requires a relatively precise pre-alignment to avoid being stuck in an undesirable local minima, the performance of our method is well suited for the problem. In some applications, the outcome of our automatic 
	alignment might be used without ICP step.   
	
	In conclusion, our method presented here solves the problem of accurately aligning partial scans using a simple, robust, and efficient approach.
	
	\bibliographystyle{eg-alpha} 
	\bibliography{ref}

\end{document}